# To learn image super-resolution, use a GAN to learn how to do image degradation first

Adrian Bulat*, Jing Yang*, Georgios Tzimiropoulos

Computer Vision Laboratory, University of Nottingham, U.K.
`{adrian.bulat,jing.yang2,yorgos.tzimiropoulos}`@nottingham.ac.uk

**Abstract.** This paper is on image and face super-resolution. The vast majority of prior work for this problem focus on how to increase the resolution of low-resolution images which are artificially generated by simple bilinear down-sampling (or in a few cases by blurring followed by down-sampling). We show that such methods fail to produce good results when applied to real-world low-resolution, low quality images. To circumvent this problem, we propose a two-stage process which firstly trains a High-to-Low Generative Adversarial Network (GAN) to learn how to degrade and downsample high-resolution images requiring, during training, only *unpaired* high and low-resolution images. Once this is achieved, the output of this network is used to train a Low-to-High GAN for image super-resolution using this time *paired* low- and high-resolution images. Our main result is that this network can be now used to effectively increase the quality of real-world low-resolution images. We have applied the proposed pipeline for the problem of face super-resolution where we report large improvement over baselines and prior work although the proposed method is potentially applicable to other object categories.

## 1 Introduction

This paper is on enhancing the resolution and quality of low-resolution, noisy, blurry, and corrupted by artefacts images. We collectively refer to all these tasks as image super-resolution. This is a challenging problem with a multitude of applications from image enhancement and editing to image recognition and object detection to name a few.

Our main focus is on the problem of super-resolving *real-world low-resolution* images *for a specific object category*. We use faces in our case noting however that the proposed method is potentially applicable to other object categories. Although there is a multitude of papers on image and face super-resolution, the large majority of them use as input low-resolution images which are artificially generated by simple bilinear down-sampling or in a few cases by blurring followed by down-sampling. On the contrary, the real-world setting has received little attention by the community. To our knowledge, this paper presents one of the very first attempts towards real-world image super-resolution. A few results produced by our system are shown in Fig. 1.

---

* Denotes equal contribution.

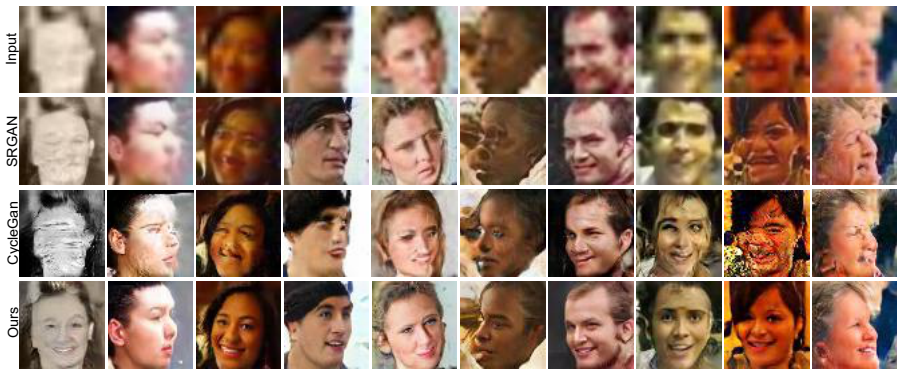

Fig. 1: Super-resolution results produced by our system on real-world low-resolution faces from Widerface [1]. Our method is compared against SRGAN [2] and CycleGan [3].

**Main idea.** There is a large list of nuisance factors which one needs to take into account when doing real-world image super-resolution, including blur (e.g. motion or defocus), compression artefacts, colour and sensor noise. These nuisance factors are usually unknown (e.g. motion blur) and sometimes hard to effectively model (e.g. the case of multiple degradations). If the true image degradation model is different from the one assumed and modeled, inevitably, this leads to poor performance during test time. To alleviate this, in this paper, rather than trying to model the image degradation process, we propose to learn it using a High-to-Low Generative Adversarial Network (GAN). Notably, the proposed network uses unpaired image data during training and hence it does not require pairs of low and high-resolution images but just two unrelated sets of low- and high-resolution images with no correspondence. Once this is achieved, we can use the High-to-Low GAN to "realistically" degrade and downsample high-resolution images and use these images as input to learn super-resolution under a "paired" image setting. The proposed architecture is shown in Fig. 2.

In summary our **contributions** are:

1. We present one of the first attempts to super-resolve real-world low-resolution images for a given object category, namely faces in this paper.
2. To this end, and inspired by [3], we propose to train a High-to-Low GAN using unpaired low- and high-resolution images which can be used to effectively simulate the image degradation process. Following this, we use the High-to-Low GAN to create paired low and high-resolution images which can be used to train a Low-to-High GAN for real-world super-resolution.
3. In recent works on image super-resolution, the $L_2$ pixel loss dominates the GAN loss which plays a refinement role in making the images look sharper.

In this work, we propose a GAN-centered approach in which the GAN loss drives the image generation process. We note that the GAN loss used plays a reciprocal role in High-to-Low and Low-to-High. In High-to-Low, it is used to contaminate the high-resolution input image with noise and artefacts coming from the Widerface dataset [1], whereas in Low-to-High it is used for denoising. In both networks, the role of the $L_2$ pixel loss is reduced to that of helping the generator preserve the face characteristics (e.g. identity, pose, expression).
4. We have applied the proposed pipeline to the problem of face super-resolution where we report large improvement over baselines and prior work on real-world, low-quality, low-resolution images from the Widerface dataset.

## 2   Closely related work

There is a very long list of image and face super-resolution papers and a detailed review of the topic is out of the scope of this section. Herein, we focus on related recent work based on Convolutional Neural Networks (CNNs).

The standard approach to super-resolution using CNNs is to use a fully supervised approach where a low-resolution (LR) image is processed by a network comprising convolutional and upsampling layers in order to produce a high-resolution (HR) image which is then matched against the original HR image using an appropriate loss function. We call this paired setting as it uses pairs of LR and corresponding HR images for training.

We emphasize that the large majority of prior work use LR images which are artificially generated by simple bilinear down-sampling of the corresponding HR images (or in a few cases by blurring followed by down-sampling). No matter the approach taken, the vast majority of image and face super-resolution methods reviewed below are based on this setting. Notably, a recent challenge on super-resolution [4] is also based on this setting. As it was recently shown in [5] and also validated in this work, this setting cannot produce good results for real-world low-resolution images.

**Image super-resolution.** Early attempts based on the aforementioned setting [6, 7] use various $L_p$ losses between the generated and the ground truth HR images for training the networks which however result in blurry super-resolved images. A notable improvement is the so-called perceptual loss [8] which applies an $L_2$ loss over feature maps calculated using another pre-trained network (e.g. VGG [9]). More advanced deep architectures for super-resolution including recursive, laplacian and dense networks have been recently proposed in [10–12].

More recently, and following the introduction of GANs [13], the authors of [2] proposed a super-resolution approach which, on top of pixel- and/or feature-based losses, it also uses a discriminator to differentiate between the generated and the original HR images which is found to produce more photo-realistic results. Notably, [14], which is an improved version of [2], won the first place in the challenge of [4]. More recently, [15] proposed a patch-based texture loss which is found to improve the reconstruction quality. Different from the aforementioned methods is [16] which does not use a GAN but proposes a pixel recursive super-resolution method which is based on PixelCNNs [17].

From the aforementioned works, our method has similar objectives to those of [5] which also targets the case of real-world image super-resolution. However, the methodology used in [5] and the one proposed in this paper are completely different. While [5] proposes to capitalize on internal image statistics to do real-world super-resolution, our method proposes to use unpaired LR and HR images to learn the image degradation process, and then use it to learn super-resolution.

**Face super-resolution.** Face super-resolution is super-resolution applied to faces. Similarly to image super-resolution, the vast majority of face super-resolution methods [18–23] are based on a paired setting for training and evaluation which is typically done on frontal datasets (e.g. CelebA [24], Helen [25], LFW [26], BioID [27]).

The method of [21] performs super-resolution and dense landmark localization in an alternating manner which is shown to improve the quality of the super-resolved faces. The authors of [19] propose a patch-based super-resolution method in which the facial regions to be enhanced are sequentially discovered using deep reinforcement learning. Rather than directly generating the HR image, the method of [20] proposes to combine CNNs with the Wavelet Transform for predicting a series of corresponding wavelet coefficients. The recent work of [22] is a GAN-based approach similar to the one proposed in [2]. In [18], a two-step decoder-encoder-decoder architecture is proposed also incorporating a spatial transformer network to undo face misalignment.

To our knowledge, the only method that reports face super-resolution results for real-world LR facial images is the very recent work of [28] which presents impressive qualitative results on more than 200 facial images taken from the Widerface dataset [1]. However, [28] is face-specific making use of facial landmarks for producing these results, rendering the approach inapplicable for other object categories for which landmarks are not available or landmark localization is not so effective. Contrary to many face super-resolution methods, the proposed pipeline is potentially applicable to other object categories.

## 3   Method

### 3.1   Overall architecture

Given a LR facial image of size $16 \times 16$, our system uses a super-resolution network, which we call Low-to-High, to super-resolve it into a HR image of $64 \times 64$. This Low-to-High network is trained with paired LR and HR facial images. One first fundamental difference between this paper and prior work on super-resolution is how the LR images are produced. In most prior work, the LR images are produced by bilinearly downsampling the corresponding (original) HR images, which completely ignores the degradation process (e.g. motion blur, compression artefacts etc.). To alleviate this, and inspired by [3], in this work, we propose to learn both degrading and downsampling a HR facial image using another network which we call High-to-Low. Notably, High-to-Low is trained using unpaired data from 2 completely different and disjoint datasets. The first of these datasets contains HR facial images from a number of face alignment datasets. The second dataset contains blurry and low quality LR facial images from Widerface.

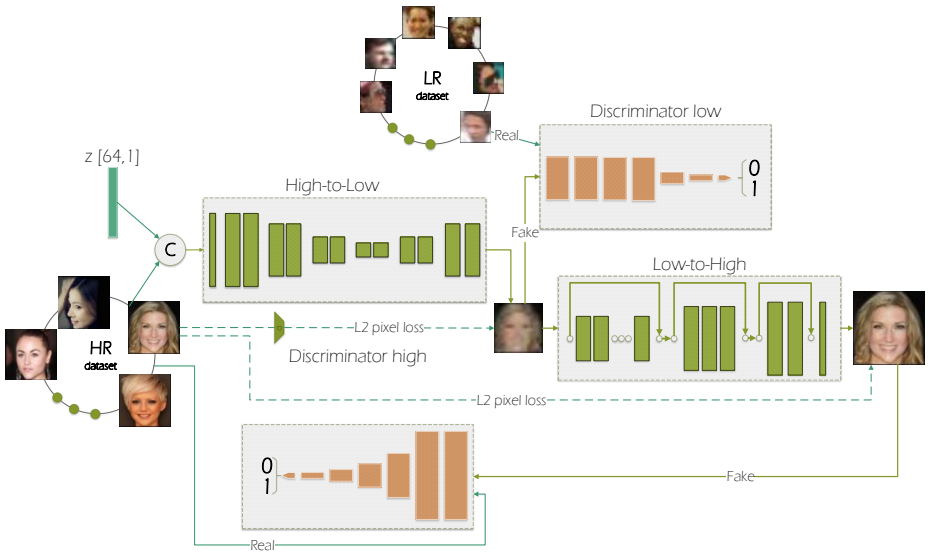

Fig. 2: Overall proposed architecture and training pipeline. See also Section 3.1.

A second fundamental difference between this paper and previous work is how the losses used to train both networks are combined with our paper putting more emphasis on the GAN over the $L_2$ pixel loss. In particular, while prior methods also use a combination of a pixel loss and GAN loss (and in some cases a feature loss), the GAN simply plays the role of making the images sharper. On the contrary, our proposed method is fully GAN-driven, with the pixel loss having the sole role of accelerating the convergence speed, especially early in the training process and helping the GAN to preserve the identity and the overall facial characteristics (e.g. pose, facial expression).

The overall architecture, which is end-to-end trainable, is shown in Fig. 2. Note that at test time only the generator part of the Low-to-High network is used. The datasets used for training and testing are described in Section 3.2. The High-to-Low and Low-to-High networks are described in detail in Sections 3.3 and 3.4, respectively. The loss functions used are detailed in Section 3.5. Finally, the training process is described in Section 3.6.

### 3.2 Datasets

This section describes the HR and LR datasets used during training and testing.

**HR dataset.** Our aim was to create a balanced dataset in terms of facial pose, hence we created a dataset of 182,866 faces by combining a series of datasets: a randomly selected subset of 60,000 faces from Celeb-A [24] (mainly frontal, occlusion-free, with good illumination), the whole AFLW [29] (more than 20,000 faces in various poses and expressions), a subset of LS3D-W [30] (faces with large variation in terms of pose, illumination, expression and occlusion), and a subset of VGGFace2 [31] (10 large pose images per identity; 9,131 identities).

**LR dataset.** We created our real-world LR dataset from Widerface [1] which is a very large scale and diverse face dataset, containing faces which are affected by a large variety of degradation and noise types. In total, we used more than 50,000 images out of which 3,000 were randomly selected and kept for testing.

### 3.3  High-to-Low

In this section, we describe the overall architecture used for the High-to-Low network. Both the generator and the discriminator are based on ResNet architectures [32, 33] using the basic block with pre-activation introduced in [33].

**High-to-Low generator.** The generator uses input images from the HR dataset. Its architecture is similar to the ones used in [2, 28] with the main difference being that the first layer takes as input the HR image concatenated with a noise vector that was projected and then reshaped using a fully connected layer in order to have the same size as one image channel. This is because the problem at hand is one-to-many, i.e. a HR image can have multiple corresponding LR ones, due to the fact that it can be affected by multiple types of noise coming from different sources and applied in different amounts and ways. We model this by concatenating the above-mentioned noise vector along with the HR image. This is similar in nature to a conditional GAN [34], in which the label is the HR image. A few visual examples illustrating the various noise types learned by the proposed network are shown in Fig. 3.

The network has an encoder-decoder structure and consists of 12 residual blocks equally distributed in 6 groups. Resolution is dropped 4 times using pooling layers, from $64 \times 64$ to $4 \times 4$ px, and then it is increased twice to $16 \times 16$ using pixel shuffle layers. The High-to-Low generator is shown in Fig. 4a.

**High-to-Low discriminator.** The discriminator, shown in Fig. 5, follows the ResNet-based architecture used in [35–37] and consists of 6 residual blocks (without batch normalization), followed by a fully connected layer. Since the input resolution of the High-to-Low discriminator is $16 \times 16$, the resolution is dropped for the last two blocks only using max-pooling.

**High-to-Low loss.** The generator and the discriminator networks of the High-to-Low network were trained with a total loss which is a combination of a GAN loss and an $L_2$ pixel loss. These are further described in Eq. 1 and detailed in Section 3.5. For the GAN loss, we used an "unpaired" training setting: in particular, we used real images from the LR dataset, i.e. real-world LR images from Widerface, therefore enforcing the output of the generator (whose input are images from the HR dataset) to be contaminated with real-world noisy artefacts. We also used an $L_2$ pixel loss between the output of the generator and the HR image after passing it through an average pooling layer (so that the image resolutions match) to enforce that the output of the generator has similar content (i.e. face identity, pose and expression) with the original HR image.

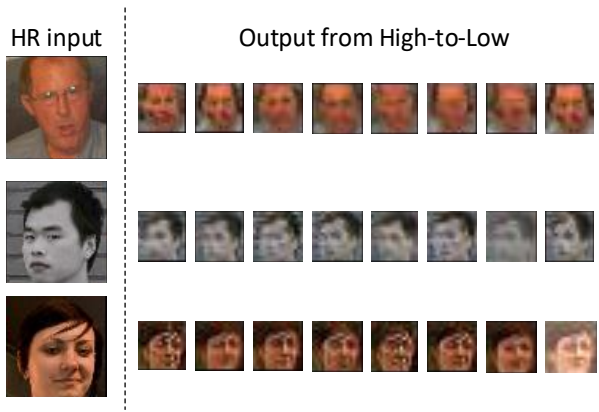

Fig. 3: Examples of different low-resolution samples produced by our High-to-Low network (described in Section 3.3) for different input noise vectors. Notice that our network can model a large variety of image degradation types, in various degrees, such as illumination, blur, colour and jpeg artefacts. Moreover, it learns what types of noise are more likely to be found given the input image type (e.g. gray-scale vs colour images). Best viewed in electronic format.

### 3.4 Low-to-High

**Low-to-High generator.** The generator accepts as input the output of the High-to-Low network. The network consists of 17 residual blocks distributed in 3 groups: 2, 3 and 12. Each group has a skip connection that connects the first and the last block within the group. Resolution is increased 4 times using bilinear interpolation, from $16 \times 16$ to $64 \times 64$ px. The generator is shown in Fig. 4b. We note that because sample diversity is already obtained at the previous stage with the help of the noise vector used in the input of the High-to-Low, we did not use an additional noise vector at this stage.

**Low-to-High discriminator.** The discriminator, shown in Fig. 5, is the same as the one used in High-to-Low, with the exception of adding two new max-pooling layers to accommodate for the increase in resolution.

**Low-to-High loss.** Similarly to High-to-Low, the generator and the discriminator networks of Low-to-High were trained with a total loss which is a combination of a GAN loss and an $L_2$ pixel loss. Note that training in this case fully follows the "paired" setting: For both losses, and for each input image, we use the corresponding image from the HR dataset. We note that, although in previous works, the GAN loss had a "secondary" role making the output image look sharper, in our case, it plays the major role for denoising the noisy input LR image. The $L_2$ pixel loss enforces content preservation.

### 3.5  Loss functions

We trained *both* High-to-Low and Low-to-High using a weighted combination of a GAN loss and an $L_2$ pixel. As mentioned earlier, a second fundamental difference between this paper and previous work is how these losses are combined. While recent works on image super-resolution also use such a combination (in many cases there is also a feature loss), in these works, the $L_2$ pixel loss dominates with the GAN loss playing a refinement role for making the images look sharper and more realistic (as the $L_2$ pixel loss is known to generate blurry images).

On the contrary, in this work, we propose a GAN-centered approach in which the GAN loss drives the image generation process. We note that the GAN loss used plays a reciprocal role in High-to-Low and Low-to-High. In High-to-Low, it is used to contaminate the HR input with noise and artefacts coming from the Widerface dataset, whereas in Low-to-High it is used for denoising. In both networks, the role of the $L_2$ pixel loss is reduced to that of helping the generator preserve the face characteristics (e.g. identity, pose, expression).

For *each network*, we used a loss defined as:

$$l = \alpha l_{pixel} + \beta l_{GAN}, \tag{1}$$

where $\alpha$ and $\beta$ are the corresponding weights and $\beta l_{GAN} > \alpha l_{pixel}$ in general.

For both networks, for the GAN loss, we made use of the recent advancements in the field and experimented with both the Improved Wasserstein [35] and the Spectral Normalization GAN [36]. From our experiments, we found that they both generated samples of similar visual quality. We note that, for our final results, we used the latter one, due to the faster training.

Following [36], we used the hinge loss defined as:

$$l_{GAN} = \mathop{\mathbb{E}}_{x \sim \mathbb{P}_r}[\min(0, -1 + D(x))] + \mathop{\mathbb{E}}_{\hat{x} \sim \mathbb{P}_g}[\min(0, -1 - D(\hat{x}))], \tag{2}$$

where $\mathbb{P}_r$ is the data distribution and $\mathbb{P}_g$ is the generator $G$ distribution defined by $\hat{x} = G(x)$. For High-to-Low, $\mathbb{P}_r$ denotes the LR dataset (i.e. the LR Widerface images), while for Low-to-High the HR dataset. See also Section 3.2.

The weights W of the discriminator D are normalized in order to satisfy the Lipschitz constraint $\sigma(W) = 1$ as follows:

$$W_{SN}(W) = W/\sigma(W). \tag{3}$$

Finally, the $L_2$ pixel loss used minimizes the $L_2$ distance between the predicted and the ground truth image and is defined as follows:

$$l_{pixel} = \frac{1}{WH} \sum_{i=1}^{W} \sum_{j=1}^{H} (F(I^{hr})_{i,j} - G_{\theta_G}(I^d)_{i,j})^2, \tag{4}$$

where $W, H$ denote the size of the generated output image and $F$ is a function that maps the corresponding original HR image $I^{hr}$ to the output resolution. For High-to-Low, this function is implemented using an average pooling layer, while, for Low-To-High, it is simply the identity function.

### 3.6 Training

To crop all facial images in a consistent manner, we ran the face detector [38] on all datasets. To further increase the diversity, we augmented the data during training by applying random image flipping, scaling, rotation and colour jittering. In order to train the Low-to-High network, we generated on-the-fly LR images, each time providing as input a different random noise vector to High-to-Low, sampled from a normal distribution, in order to simulate a large variety of image degradation types. Both the High-to-Low and Low-to-High networks were trained for 200 epochs (about 570,000 generator updates), with an update ratio 5:1 between the discriminator and the generator. In the end, we fine-tuned them together for another 2,000 generator updates. The learning rate was kept to $1e-4$ for the entire duration of the training process. We used $\alpha = 1$ and $\beta = 0.05$ in Eq. 1. All of our models were trained using PyTorch [39] and optimized with Adam [40] ($\beta_1 = 0$ and $\beta_2 = 0.9$).

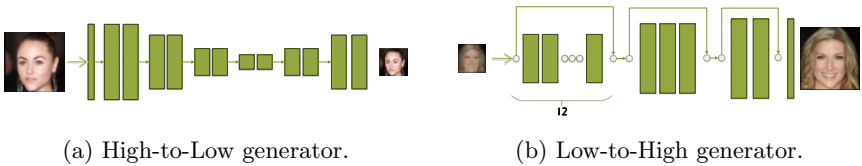

(a) High-to-Low generator.   (b) Low-to-High generator.

Fig. 4: The generator architecture used for the (a) High-to-Low and (b) Low-to-High networks. The residual block used is shown in Fig. 6b.

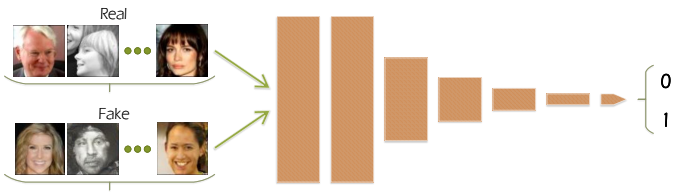

Fig. 5: The discriminator architecture used for both High-to-Low and Low-to-High networks. Note that, for High-to-Low, the first two max-pooling layers are omitted since the input resolution is $16 \times 16$. The residual block used is shown in Fig. 6a.

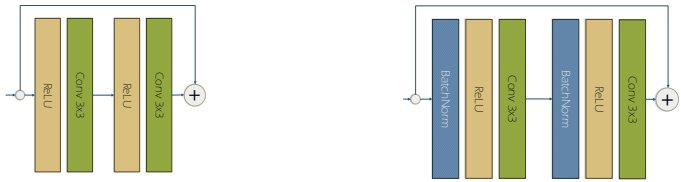

(a) Residual block with pre-activation and without batchnorm.

(b) Residual block with pre-activation and batch normalization as defined in [33].

Fig. 6: The residual blocks used for the discriminator (a) and the generator (b).

## 4  Results

In this section, we evaluate the performance of our system, and compare it with that of a few interesting variants and related state-of-the-art methods. Our main results are on the 3,000 images of our LR test set which contains images from the Widerface dataset. For this experiment, and because there are no corresponding ground-truth HR images, besides visual results, we numerically assess the quality of the generated samples using the Fréchet Inception Distance (FID) [41]. Finally, for completeness, we also provide PSNR results on 1,000 test images from the LS3D-W dataset using bilinearly downsampled images as input. This is the standard super-resolution experimental setting used in prior work.

### 4.1  Methods compared

**Other variants.** Alongside the proposed method presented in Section 3, we also evaluate the performance of a series of interesting variants, all of which are detailed as follows:

1. Low-to-High-trained-on-bilinear: this is the Low-to-High network of Section 3.4 trained on images that were bilinearly downsampled. The network is trained with the loss of Eq. 1.
2. Low-to-High-trained-on-bilinear-blur: this is the Low-to-High network of Section 3.4 trained on bilinearly downsampled images after being blurred with a random blur kernel with a kernel size varying from 2 to 6 px. The network is trained with the loss of Eq. 1.
3. Low-to-High+Low-to-High-pixel-loss: this is the Low-to-High network of Section 3.4 that uses the High-to-Low network of Section 3.3 to generate the LR training samples. The Low-to-High network is trained *only using the $L_2$ pixel loss* in Eq. 4.
4. Low-to-High+High-to-Low-pixel-and-gan-loss: this is the Low-to-High network of Section 3.4 that uses the High-to-Low network of Section 3.3 to generate the LR training samples. The network is trained using the loss defined in Eq. 1. *This is the full implementation of the proposed method.*

**State-of-the-art.** Our method is compared both numerically and qualitatively with 5 related state-of-the-art methods: 1 image super-resolution method, namely SRGAN [2], 2 face super-resolution methods, namely Wavelet-SRNet [20] and FSRNet [42], 1 unpaired image translation method, namely CycleGan [3], and 1 deblurring method, namely DeepDeblur [43]. SRGAN and Wavelet-SRNet were trained on our training set using pairs of bilinearly downsampled - HR images. FSRNet provides only testing code (trained on their dataset using pairs of bilinearly downsampled - HR images). CycleGan was trained similarly to our method. Finally, for DeepDeblur we had 2 options: either use the pre-trained model trained on their data (pairs of blurred - clear images) or re-train it on our training set using pairs of bilinearly downsampled - HR images. The latter option would make it very similar to SRGAN, hence we used the former option.

### 4.2   Super-resolution results

Quantitative results on our LR test set in terms of FID are shown in Table 1. Qualitative results for several images are shown in Figs. 7 and 8. The visual results for *all 3,000* test images can be found in the supplementary material. Moreover, we provide PSNR results on LS3D-W in Table 1. Our method clearly outperforms all other variants and methods considered both numerically (in terms of FID) and (more importantly) visually.

**Comparison with other variants.** As expected, the Low-to-High trained on bilinearly downsampled images (Low-to-High-trained-on-bilinear) does not perform well and neither does the Low-to-High trained on bilinearly downsampled images after blurring them with various blur kernels (Low-to-High-trained-on-bilinear-blur). Overall, the results obtained by these methods are both noisy and blurry. Because of this, we propose to learn the noise distribution from Widerface images using the High-to-Low network. Directly training however such network using an $L_2$ pixel loss does not work well (Low-to-High+Low-to-High-pixel-loss). We conclude that the $L_2$ loss alone is not able to denoise the input and produce good results. However, once the GAN loss (proposed method) is added, the network is successfully able to both (a) produce high quality samples and (b) denoise the images for most of the cases.

In addition to the above results, we also tried to quantify how close the generated by the High-to-Low network images resemble the original LR images from Widerface. To this end, their FID was found equal to 15.27 while the FID between bilinearly downsampled and original LR images was found equal to 23.15. This result clearly illustrates the effectiveness of the proposed High-to-Low network in producing images that faithfully represent real-world degradations.

**Comparison with the state-of-the-art.** In terms of FID, our method largely outperforms all other methods. Moreover, from Figs. 7 and 8, we observe that our method produces the most appealing visual results. Both results show that, in contrary to all other methods considered, our High-to-Low network can model the image degradation in real LR datasets. Although CycleGan also achieves relatively low FID, from Figs. 7 and 8, it can be observed that visually the produced results are of low quality. This is because the cycle consistency loss emphasizes too much on pixel-level similarity, creating a lot of noise artifacts in the final output.

Finally, regarding our experiment on LS3D-W, as all other methods were trained on pairs of bilinearly downsampled and original HR images, they have an advantage and outperform our method. Our method however (trained using the output of High-to-Low) provides competitive results (PSNR $\approx$ 20 dB).

| Method | FID | PSNR |
|---|---|---|
|  | LR test set | LS3D-W |
| SRGAN [2] | 104.80 | 23.19 |
| CycleGan [3] | 19.01 | 16.10 |
| DeepDeblur [43] | 294.96 | 19.62 |
| Wavelet-SRNet [20] | 149.46 | **23.98** |
| FSRNet [42] | 157.29 | 19.45 |
| Low-to-High (trained on bilinear) | 85.59 | 23.50 |
| Low-to-High (trained on blur + bilinear) | 84.68 | 22.87 |
| High-to-Low+Low-to-High (pixel loss only) | 87.91 | 23.22 |
| **Ours** | **14.89** | 19.30 |

Table 1: (a) FID-based performance on our real-world LR test set. Lower is better. (b) PSNR results on LS3D-W (the input LR images are bilinearly downsampled images).

### 4.3  Failure cases

By no-means we claim that the proposed method solves the real-world image and face super-resolution problem. We show several failure cases of our method in Fig. 9. We can group failures into two groups: the first one contains cases of complete failures where the produced image does not resemble a face. For many of these cases, we note that the input does not resemble a face either. Examples of these cases are shown in the first two rows of Fig. 9. The second group contains cases that the produced super-resolved face is distorted. These are mostly cases of extreme blur, occlusion and large pose. Examples of these cases are shown in the last two rows of Fig. 9. We need to emphasize here that many of the large pose facial images of the HR dataset used for training are synthetically warped into these poses (see [30]), and this is expected to have some negative impact on performance. In total, we found that the percentage of fail cases in our test set is about 10%.

## 5  Conclusions

We presented a method for image and face super-resolution which does not assume as input artificially generated LR images but aims to produce good results when applied to real-world, LR, low quality images. To this end, we proposed a two-stage process which firstly uses a High-to-Low network to learn how to downgrade high-resolution images requiring only *unpaired* high- and low-resolution images and uses the output of this network to train a Low-to-High network for image super-resolution. We showed that our pipeline can be used to effectively increase the quality of real-world LR images. We reported large improvement over baselines and prior work.

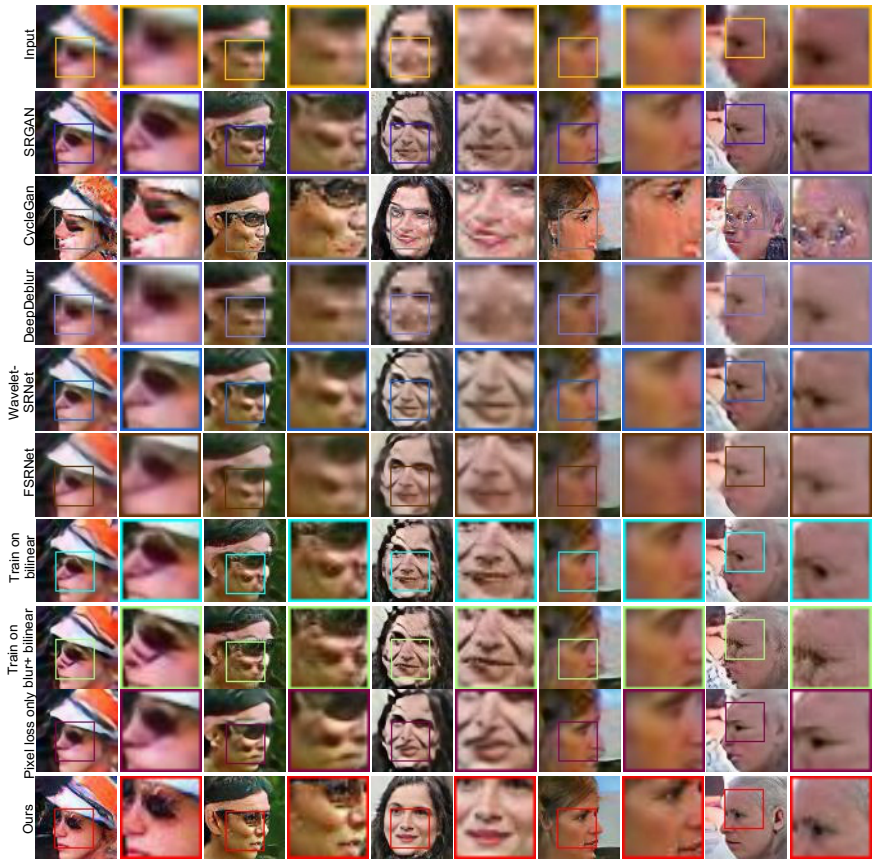

Fig. 7: Detailed qualitative results on our LR test set from Widerface. The methods compared are described in Section 4.1.

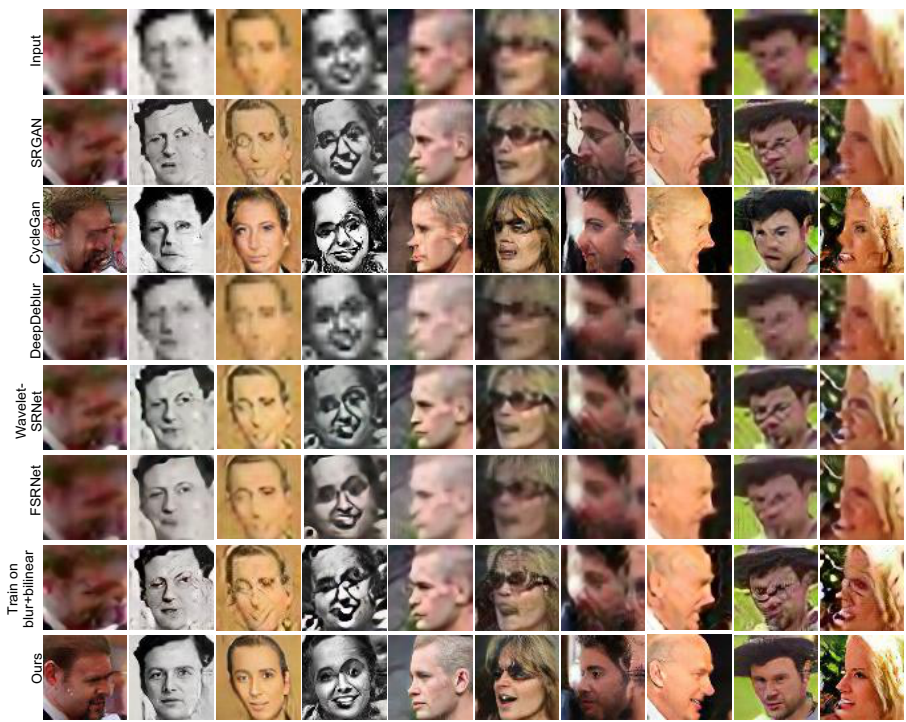

Fig. 8: Additional qualitative results on our LR test set from Widerface. The methods compared are described in Section 4.1.

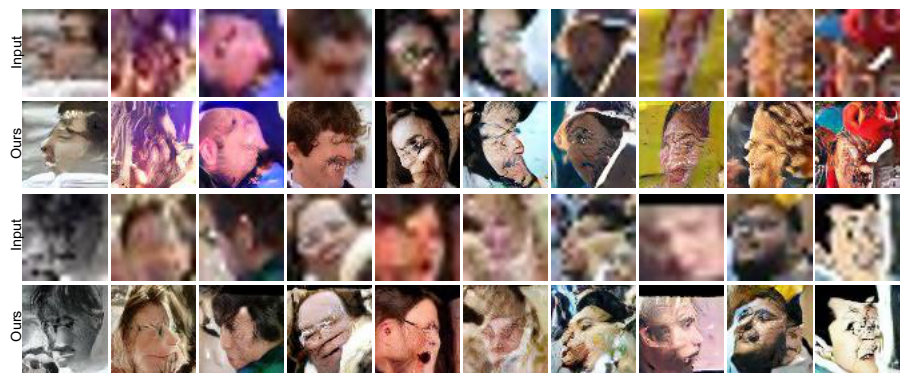

Fig. 9: Examples of failure cases. The input images are shown in the first and third row while the output images produced by our method in the second and fourth row, respectively. The images shown in the second row do not resemble a face. The images in the fourth row do resemble a face but they are heavily distorted.